\def\year{2019}\relax
\documentclass[letterpaper]{article} %DO NOT CHANGE THIS
\usepackage{aaai19}  %Required
\usepackage{times}  %Required
\usepackage{helvet}  %Required
\usepackage{courier}  %Required
\usepackage{url}  %Required
\usepackage{graphicx}  %Required
\usepackage{color}
\usepackage{latexsym}
\usepackage{graphicx}
\usepackage{subfig}
\newif\ifcomment\commenttrue
\usepackage{amssymb}
\usepackage{amsmath}

\usepackage{booktabs}

\usepackage{url}
\usepackage{footnote}
\makesavenoteenv{table}
\usepackage{algorithm}
\usepackage{algpseudocode}
\usepackage{placeins}

\newtheorem{theorem}{Theorem}
\newtheorem{lemma}{Lemma}%[section]
\newtheorem{assumption}{Assumption}

\makeatletter
\def\@biblabel#1{}
\makeatother

\title{Block Belief Propagation for Parameter Learning in Markov Random Fields}

\author{You Lu\\ Department of Computer Science \\
	Virginia Tech \\
	Blacksburg, VA\\
	\tt you.lu@vt.edu
\And
	Zhiyuan Liu \\ Department of Computer Science \\
	University of Colorado Boulder \\
	Boulder, CO \\
	\tt zhiyuan.liu@colorado.edu
\And
	Bert Huang\\ Department of Computer Science \\
	Virginia Tech \\
	Blacksburg, VA \\
	{\tt bhuang@vt.edu}
}

\date{}

\begin{document}
	
	\maketitle
	
	\begin{abstract}
		
		Traditional learning methods for training Markov random fields require doing inference over all variables to compute the likelihood gradient. The iteration complexity for those methods therefore scales with the size of the graphical models. In this paper, we propose \emph{block belief propagation learning} (BBPL), which uses block-coordinate updates of approximate marginals to compute approximate gradients, removing the need to compute inference on the entire graphical model. Thus, the iteration complexity of BBPL does not scale with the size of the graphs. We prove that the method converges to the same solution as that obtained by using full inference per iteration, despite these approximations, and we empirically demonstrate its scalability improvements over standard training methods. 
	\end{abstract}

\section{Introduction}
\label{sec:introduction}

Markov random fields (MRFs) and conditional random fields (CRFs) are powerful classes of models for learning and inference of factored probability distributions~\cite{koller2009probabilistic,wainwright2008graphical}. They have been widely used in tasks such as structured prediction~\cite{taskar2004max} and computer vision~\cite{nowozin2011structured}. Traditional training methods for MRFs learn by maximizing an approximate maximum likelihood. Many such methods use variational inference to approximate the crucial partition function. 

With MRFs, the gradient of the log likelihood with respect to model parameters is the marginal vector. With CRFs, it is the expected feature vector. These identities suggest that each iteration of optimization must involve computation of the full marginal vector, containing the estimated marginal probabilities of all variables and all dependent groups of variables. In some applications, the number of variables can be massive, making traditional, full-inference learning too expensive in practice. This problem limits the application of MRFs in modern data science tasks.

In this paper, we propose \emph{block belief propagation learning} (BBPL), which alleviates the cost of learning by computing approximate gradients with inference over only a small block of variables at a time. BBPL first separates the Markov network into several small blocks. At each iteration of learning, it selects a block and computes its marginals. It approximates the gradient with a mix of the updated and the previous marginals, and it updates the parameters of interest with this gradient. 

\subsection{Related Work} 

%BBPL builds on existing knowledge about scaling learning and inference with MRFs. Other approaches have been developed to improve convergence properties of inference methods, to change the learning objective to reduce the cost of inference during learning, or to reduce the cost of inference during learning by running fewer iterations of message passing.

Many methods have been developed to learn MRFs. In this section, we cover only the BP-based methods.

Mean-field variational inference and belief propagation (BP) approximate the partition function with non-convex entropies, which break the convexity of the original partition function. In contrast, convex BP~\cite{globerson2007approximate,heskes2006convexity,schwing2011distributed,wainwright2005new,wainwright2006estimating} provides a strongly convex upper bound for the partition function. This strong convexity has also been theoretically shown to be beneficial for learning~\cite{london2015benefits}. Thus, our BBPL method uses convex BP to approximate the partition function.

Regarding inference, some methods are developed to accelerate computations of the beliefs and messages. Stochastic BP~\cite{noorshams2013stochastic} updates only one dimension of the messages at each inference iteration, so its iteration complexity is much lower than traditional BP. Distributed BP~\cite{schwing2011distributed,yin2014scalable} distributes and parallelizes the computation of beliefs and messages on a cluster of machines to reduce inference time. Sparse-matrix BP~\cite{bixler2018sparse} uses
sparse-matrix products to represent the message-passing indexing, so that it can be implemented on modern hardware. However, to learn MRF parameters, we need to run these inference algorithms for many iterations on the whole network until convergence at each parameter update. Thus, these methods are still impacted by the network size.

%Many methods have been developed to efficiently infer MRF marginals, i.e., beliefs, or learn its parameters. In this section, we cover only the BP-based methods. In Stochastic BP~\cite{noorshams2013stochastic} updates only one dimension of the messages at each inference iteration, so its iteration complexity is much lower than traditional BP. Distributed BP~\cite{schwing2011distributed,yin2014scalable} distribute and parallelize the computation of beliefs and messages to a cluster of machines to reduce the inference time. 

Regarding learning, many learning frameworks have been proposed to efficiently learn MRF parameters. Some approaches use neural networks to directly estimate the messages~\cite{lin2015deeply,ross2011learning}. Methods that truncate message passing~\cite{domke2011parameter,domke2013learning,stoyanov2011empirical}  redefine the loss function in terms of the approximate inference results obtained after message passing is truncated to a small number of iterations. These methods can be inefficient because they require running inference on the full network for a fixed number of iterations or until convergence. Moreover, they yield non-convex objectives without convergence guarantees. 

Some methods restrict the complexity of inference on a subnetwork at each parameter update. Lifted BP~\cite{kersting2009counting,singla2008lifted} makes use of the symmetry structure of the network to group nodes into super-nodes, and then it runs a modified BP on this lifted network. A similar approach~\cite{ahmadi2012lifted} uses a lifted online training framework that combines the advantages of lifted BP and stochastic gradient methods to train Markov logic networks~\cite{richardson2006markov}. These lifting approaches rely on symmetries in relational structure---which may not be present in many applications---to reduce computational costs. They also require extra algorithms to construct the lifted network, which increase the difficulty of implementation. Piecewise training separates a network into several possibly overlapping sets of pieces, and then uses the piecewise pseudo-likelihood as the objective function to train the model~\cite{sutton08piecewise}. Decomposed learning~\cite{samdani2012efficient} uses a similar approach to train structured support vector machines. However, these methods need to modify the structure of the network by decomposing it, thus changing the objective function. 

Finally, inner-dual learning methods~\cite{bach2015paired,hazan2010primal,hazan2016blending,meshi2010learning,taskar2005learning} interleave parameter optimization and inference to avoid repeated inferences during training. These methods are fast and convergent in many settings. However, in other settings, key bottlenecks are not fully alleviated, since each partial inference still runs on the whole network, causing the iteration cost to scale with the network size. 

\subsection{Contributions}

In contrast to many existing approaches, our proposed method uses the same convex inference objective as traditional learning, providing the same benefits from learning with a strongly convex objective. BBPL thus scales well to large networks because it runs convex BP on a block of variables, fixing the variables in other blocks. This block update guarantees that the iteration complexity does not increase with the network size. BBPL preserves communication across the blocks, allowing it to optimize the original objective that computes inference over the entire network at convergence.
The update rules of BBPL are similar in form to full BP learning, which makes it as easy to implement as traditional MRF or CRF training. Finally, we theoretically prove that BBPL converges to the same solution under mild conditions defined by the inference assumption. Our experiments empirically show that BBPL does converge to the same optimum as full BP. 

\section{Background}
\label{sec:background}

In this section, we introduce notation and background knowledge directly related to our work.

\subsection{Convex Belief Propagation for MRFs}

Let $x = [x_1, . . . , x_n]$ be a discrete random vector taking values in $\mathcal{X} = \mathcal{X}_1 \times \dots \times \mathcal{X}_n$, and let $G = (V, E)$ be the corresponding undirected graph, with the vertex set $V = \{1,...,n\}$ and edge set $E \subset V \times V$. Potential functions $\theta_s: \mathcal{X}_s \to \mathcal{R}$ and $\theta_{uv}: \mathcal{X}_u \times \mathcal{X}_v  \to \mathcal{R}$ are differentiable functions with parameters we want to learn. The probability density function of a pairwise Markov random field~\cite{wainwright2008graphical} can be written as
\begin{eqnarray*}
	p(x| \theta) = \exp\{\sum_{s \in V} \theta_s(x_s) + \sum_{(u,v) \in E} \theta_{uv}(x_u,x_v) - A(\theta)\}.
\end{eqnarray*}

The log partition function
\begin{equation}
A(\theta) = \log \sum_{X} \exp\{\sum_{s \in V} \theta_s(x_s) + \sum_{(u,v) \in E} \theta_{uv}(x_u,x_v) \}
\label{eq:A}
\end{equation}
is intractable in most situations when $G$ is not a tree. One can approximate the log partition function with a convex upper bound:
\begin{equation}
B(\theta) = \max_{\tau \in \mathbb{L}(G)}\{\langle \theta, \tau \rangle - B^*(\tau) \},
\label{eq:B}
\end{equation}
where 
\begin{eqnarray*}
	\theta &=& \{\theta_s| s\in V\} \cup \{\theta_{uv}| (u,v) \in E\}, \\
	\tau &=& \{ \tau_s | s \in V  \} \cup \{ \tau_{uv} | (u,v) \in E \}, \\	
	\mathbb{L}(G) &:=& \{\tau \in \mathbb{R}_+^d | \sum_{x_s}\tau_{s}(x_s) = 1, \\
	&& \sum_{x_v}\tau_{uv}(x_u,x_v)= \tau_{u}(x_u)\}.
\end{eqnarray*}

The vector $\tau$ is called the pseudo-marginal or belief vector. Specifically, $\tau_s$ is the unary belief of vertex $s$, and $\tau_{uv}$ is the pairwise belief of edge $(u,v)$. The \emph{local marginal polytope} $\mathbb{L}$ restricts the unary beliefs to be consistent with their connected pairwise beliefs. We consider a variant of belief propagation where $B^*(\tau)$ is strongly convex and has the following form:
%\begin{eqnarray*}
%B^*(\tau) &=&  -\sum_{s \in V}\rho_s (\sum_{x_s} \tau_s(x_s)\log \tau_s(x_s)) \\ 
%&+& \sum_{(u,v) \in E}\rho_{uv}  \hspace{-0.03in} \sum_{x_u,x_v} \tau_{uv}(x_u,x_v) \log %\frac{\tau_{uv}(x_u,x_v)}{\tau_{x_u}(x_u)\tau_{x_v}(x_v)},
%\end{eqnarray*} 

\begin{eqnarray*}
	B^*(\tau) &=& \sum_{s \in V}\rho_s H( \tau_s) + \sum_{(u,v) \in E}\rho_{uv} H(\tau_{uv}),
\end{eqnarray*} 
where $\rho_s$ and $\rho_{uv}$ are parameters known as counting numbers, and $H(.)$ is the entropy.

Equation~\ref{eq:B} can be solved via convex BP~\cite{meshi2009convexifying,yedidia2005constructing}. Let $\lambda_{uv}$ be the message from vertex $u$ to vertex $v$. The update rules of messages and beliefs are as follows:
\begin{eqnarray}
\lambda_{uv} = \rho_{uv} \log \sum_{u}\exp\{ \frac{1}{\rho_{uv}}(\theta_{uv}  - \lambda_{vu})+ \log \tau_u  \},
\label{update_lambda}
\end{eqnarray}
where
\begin{equation}
\tau_u \propto \exp\{\frac{1}{\rho_{u}}(\theta_u + \sum_{v \in N(u)} \lambda_{vu})\},
\label{update_tau}
\end{equation}
and
\begin{eqnarray}
\tau_{uv} &\propto& \exp\{\frac{1}{\rho_{uv}}(\theta_{uv}-\lambda_{uv}-\lambda_{vu}) + \log \tau_u \tau_v\}.
\label{update_pair_tau}
\end{eqnarray}

Other forms of convex BP, such as tree-reweighted BP~\cite{wainwright2005new}, can also be used in our approach. Convex BP does not always converge on loopy networks. However, under some mild conditions, it is guaranteed to converge and can be a good approximation for general networks~\cite{roosta2008convergence}.

\subsection{Learning Parameters of MRFs}
In this subsection, we introduce traditional training methods for fitting MRFs to a dataset via a combination of BP and a gradient-based optimization. The learning algorithm is given a dataset with $N$ data points, i.e., $w_1,...,w_N$. It then learns $\theta$ by minimizing the negative log-likelihood:
\begin{eqnarray}
L(\theta) &=& -\frac{1}{N}\sum_{n} \log p(w_n|\theta) \nonumber \\
&=& -\frac{1}{N}\sum_{n} \theta^T w_n+ A(\theta) \nonumber\\
&\approx& -\theta^T \bar{w}+ B(\theta) \nonumber\\
&=& -\theta^T \bar{w} + \max_{\tau \in \mathbb{L}(G)}\{\langle \theta, \tau \rangle - B^*(\tau) \},
\label{eq:L}
\end{eqnarray}
where $\bar{w} = 1/N \sum_n w_n$.

Using $B(\theta)$ as the tractable approximation of $A(\theta)$, the traditional learning approach is to minimize $L(\theta)$ using gradient-based methods. Let $\theta_t$ be the parameter vector at iteration $t$, and let $\tau_t^*$ be the optimized $\tau$ corresponding to $\theta_t$. 
Then gradient learning is done by iterating
\begin{equation}
\theta_{t+1} = \theta_t - \alpha_t \nabla_\theta L(\theta_t),
\label{eq:theta}
\end{equation}
where 
\begin{equation}
\nabla_\theta L(\theta_t) = -\bar{w} + \tau^*_t,
\label{eq:true_gradient}
\end{equation}
and $\alpha_t$ is the learning rate. The traditional parameter learning process is described in Algorithm~\ref{BP}. 

\begin{algorithm}[htp]
	\caption{Parameter learning with full convex BP}
	\begin{algorithmic}[1]
		\State Initialize $\theta_0$.
		\State While $\theta$ has not converged
		\State \quad \quad While $\lambda$ has not converged
		\State \quad \quad \quad \quad Update $\tau$ with Equation~\ref{update_tau}.
		\State \quad \quad \quad \quad Update $\lambda$ with Equation~\ref{update_lambda}.
		\State \quad \quad end
		\State \quad \quad $\nabla L(\theta_t) = -\bar{w} + \tau^*_{t}$
		\State \quad \quad $\theta_{t+1} = \theta_{t} - \alpha_t \nabla L(\theta_t)$
		\State end
	\end{algorithmic}
	\label{BP}
\end{algorithm}

Traditional parameter learning requires running inference on each full training example per gradient update, so we refer to it as \emph{full BP learning}. These full inferences cause it to suffer scalability limitations when the training data includes large MRFs. The goal of our contributions is to circumvent these limitations.

\section{Block Belief Propagation Learning}
\label{sec:partialBP}

In this section, we introduce our proposed method, which has a much lower iteration complexity than full BP learning and is guaranteed to converge to the same optimum as full convex BP learning.

\subsection{Algorithm Description}

The full BP learning method in Algorithm~\ref{BP} does not scale well to large networks. It needs to perform inference on the full network at each iteration, i.e., Step 2 to Step 5 in Algorithm~\ref{BP}. The iteration complexity depends on the size of network. When the network is large---e.g., a network with at least tens of thousands of nodes---the dimensions of $\tau$ and $\lambda$ are also large. Updating all their dimensions in each gradient step creates a significant computational inefficiency.

To address this problem, we develop \emph{block belief propagation learning} (BBPL). BBPL only needs to do inference on a subnetwork at each iteration, which means that, for templated models, the iteration complexity of BBPL does not depend on the size of the network. For non-templated models, the iteration complexity has a greatly reduced dependency on the network size. Moreover, we can prove the convergence of BBPL, guaranteeing that it will converge to the same optimum as full BP learning. Finally, BBPL's update rules are analogous to those of full BP learning, so it is as easy to implement as full BP learning. 

BBPL first separates the vertex set $V$ into $D$ subsets $V_1, ..., V_D$. Then it separates the edge set $E$  into $D$ subsets of edges incident on the corresponding vertex subsets, i.e., $E_i = \{(u,v)| (u,v) \in E \quad \text{and} \quad (u \in V_i \quad \text{or} \quad v \in V_i) \}$. Thus, we have that $V = V_1 \cup V_2 \cup...\cup V_D$, and $E = E_1 \cup E_2 \cup...\cup E_D$. For shorthand, let $F_i = (E_i, V_i)$ denote the $i$th subnetwork.

At iteration $t$, BBPL selects a subgraph $F_i$ and only updates each $\tau_u$ if $u \in V_i$, and $\tau_{uv}$ and only $\lambda_{uv}$ if $(u,v) \in E_i$. It updates this block of messages and beliefs via belief propagation---i.e., Equations~\ref{update_lambda}, \ref{update_tau}, and \ref{update_pair_tau}---holding all other messages and beliefs fixed until convergence. Finally, BBPL uses the updated block of beliefs to update $\theta$ by computing an approximate gradient. Our empirical results show that either selecting subgraphs randomly or sequentially can lead to convergence, but sequentially selecting subgraphs can make the algorithm converge faster.

With a slight abuse of notation, let $F_t$ be the subnetwork selected at iteration $t$. Let  $\tau_{t}^{(F_t)}$ and $\lambda_{t}^{(F_t)}$ be the sub-matrices of $\tau_t$ and $\lambda_t$, respectively, that correspond to $F_t$. Let $U_{F_t}$ be a projection matrix that projects the parameter from $\mathbb{R}^{|F_t|}$ to $\mathbb{R}^d$. The update rules for $\tau_t$ and $\lambda_t$ are
\begin{equation}
\tau_{t} = \tau_{t-1} + U_{F_t}(\tau_{t}^{(F_t)} - \tau_{t-1}^{(F_t)}),
\label{eq:sub_tau}
\end{equation}
and
\begin{equation}
\lambda_{t} = \lambda_{t-1} + U_{F_t}(\lambda_{t}^{(F_t)} - \lambda_{t-1}^{(F_t)}).
\label{eq:sub_lambda}
\end{equation}

After computing $\tau_t$ and $\lambda_t$, BBPL updates the parameters using the approximate gradient:
\begin{equation}
g(\theta_t) = -\bar{w} + \tau_{t}.
\label{eq:appro_gradient}
\end{equation}
Note that the only difference between $g(\theta_t)$ and $g(\theta_{t-1})$ is the term $\tau_{t}^{(F_t)} - \tau_{t-1}^{(F_t)}$. Thus, given $g(\theta_{t-1})$, we can efficiently update our gradient estimate with
\begin{equation*}
g(\theta_t) = g(\theta_{t-1}) + U_{F_t}(\tau_{t}^{(F_t)} - \tau_{t-1}^{(F_t)}).
\end{equation*}
This equation implies that computation of the gradient also does not depend on the size of the network. Changing the gradient at the entries where the block marginal update was performed can be done in place. Thus, the complexity of the whole inference process and gradient computation depends only on the size of the subnetwork.
The only step that requires time complexity that scales with the network size is the actual parameter update using the gradient, which we will later alleviate when using templated models.
The complete algorithm is listed in Algorithm~\ref{PBP}. 

\begin{algorithm}[!htp]
	\caption{Parameter estimation with block BP}
	\begin{algorithmic}[1]
		\State Separate $g$ into $M$ subnetwork, i.e., $F_1,...,F_M$.
		\State While $\theta$ has not converged
		\State \quad \quad Select a subgraph $F_t$.
		\State \quad \quad While $\lambda^{F_t}$ has not converged
		\State \quad \quad \quad \quad Update $\tau_t^{F_t}$ with Equation~\ref{update_tau}.
		\State \quad \quad \quad \quad Update $\lambda_t^{F_t}$ with Equation~\ref{update_lambda}.
		\State \quad \quad end
		%		\State \quad \quad $\tau_{t} = \tau_{t-1} + U_{F_t}(\tau_{t}^{(F_t)} - \tau_{t-1}^{(F_t)})$.
		%		\State \quad \quad $\lambda_{t} = \lambda_{t-1} + U_{F_t}(\lambda_{t}^{(F_t)} - \lambda_{t-1}^{(F_t)})$.
		\State \quad \quad $g(\theta_t) = g(\theta_{t-1}) + U_{F_t}(\tau_{t}^{(F_t)} - \tau_{t-1}^{(F_t)}).$
		\State \quad \quad $\theta_{(t+1)} = \theta_{(t)} - \alpha_t g(\theta_t)$.
		\State end
	\end{algorithmic}
	\label{PBP}
\end{algorithm}
\FloatBarrier

\subsection{Convergence Analysis}

In this subsection, we theoretically prove the convergence of BPPL. We first rewrite the learning problem as follows:

\begin{equation}
\min_{\theta} \max_{\tau \in \mathbb{L}(G)} -\theta^T \bar{w} + \theta^T \tau - B^*(\tau).
\label{eq:obj}
\end{equation}

Equation~\ref{eq:obj} is a convex-concave saddle-point problem. We use $(\theta^*, \tau^*)$ to represent its saddle point. Its corresponding primal problem is 
\begin{equation}
\min_{\theta} L(\theta) = -\theta^T \bar{w} + B(\theta),
\label{eq:primal}
\end{equation}
where $B(\theta)$ is defined in Equation~\ref{eq:B}. Equation~\ref{eq:obj} and Equation~\ref{eq:primal} have the same optimal $\theta$. 

Thus, we can prove the convergence of BBPL following the general framework for proving the convergence of saddle-point optimizations~\cite{du2018linear}. We prove that under a mild assumption, i.e., Assumption~\ref{assumption:contraction}, BBPL has a linear convergence rate. 

\begin{assumption}
	When $\theta$ has not converged, the block coordinate update, i.e., Equation~\ref{eq:sub_tau}, satisfies the following inequality:
	\begin{equation*}
	||\tau_{t+1} - \tau^*_{t+1}|| \le (1-c) ||\tau_t - \tau^*_{t+1}||,
	\end{equation*}
	where $0 < c<1$.
	\label{assumption:contraction}
\end{assumption}

Informally, we can interpret Assumption~\ref{assumption:contraction} to mean that, at iteration $t+1$, the vector $\tau_{t+1}$ with new block $\tau_{t+1}^{(F_t)}$ will get closer to the optimum $\tau^*_{t+1}$. This assumption is easy to satisfy when $\theta$ has not converged. Since the block $\tau_{t+1}^{(F_t)}$ of $\tau_{t+1}$ is updated with respect to $\theta_{t+1}$, and $\tau_{t+1}^*$ is the true optimum with respect to $\theta_{t+1}$, $\tau_{t+1}$ should be closer to $\tau_{t+1}^*$. When $\theta$ converges, block BP becomes a block coordinate update method. Based on claims shown by~\citeauthor{schwing2011distributed} (\citeyear{schwing2011distributed}), $\tau_t$ converges to $\tau^*$.

The following theorem establishes the linear convergence guarantee of Algorithm~\ref{PBP}.

\begin{theorem}
	The primal of BBPL, i.e., Equation~\ref{eq:primal}, is $\beta$-strongly convex and $\eta$-smooth. When we use BBPL to learn the parameters, suppose BBPL satisfies Assumption~\ref{assumption:contraction}. Define the Lyapunov function
	\begin{equation*}
	P_t = ||\theta_t - \theta^*|| + \gamma ||\tau_t - \tau_t^*||.
	\end{equation*} 
	When $0 < \alpha_t \le \min\{\frac{c \beta}{2\eta^2 + \eta \beta + \beta^2}, \frac{2}{\eta + \beta}\}$, and $\gamma = \frac{\beta}{2(1-c)\eta^2}$, we have 
	\begin{equation*}
	P_{t+1} \le  (1 - \delta)P_t,
	\end{equation*}
	where $0 < \delta < 1$. This bound implies that $\lim_{t \to \infty}P_t = 0$, and $\theta_t$ will converge linearly to the optimum $\theta^*$.
	\label{theorem:converge}
\end{theorem}

\textit{Proof sketch.} The proof is based on the proof framework detailed by~\citeauthor{du2018linear} (\citeyear{du2018linear}). The proof can be divided to three steps. First, we bound the decrease of $||\theta_t - \theta^*||$. Second, we bound the decrease of $||\tau_t - \tau_t^*||$. Third, we prove that $P_{t+1} \le (1-\delta) P_t$. The complete proof is in the
%
% long version of this paper \cite{lu2018block}. % for AAAI version
%
appendix. % for arxiv version
\hfill $\square$

\subsection{Generalization to Templated or Conditional Models}

So far, we have described the BBPL approach in the setting where there is a separate parameter for every entry in the marginal vector. In templated or conditional models, the parameters can be shared across multiple entries. We describe here how BBPL generalizes to such models. For conditional models, we use MRFs to infer the probability of output variables $Y$ given input variables $X$, i.e., $\Pr(Y | X)$. A standard modeling technique for this task is to encode the joint states of the input and output variables with some feature vectors, so the parameters are weights for these features.
We are given a dataset of fully observed input-output pairs $S = \{(M_i, y_i)\}_{i=1}^{N}$, where $M_i \in \mathbb{R}^{K \times d}$ is the feature matrix, and $y_i \in \mathbb{R}^d$ is the label vector (i.e., the one-hot encoding of the ground-truth variable states). The negative log-likelihood of a conditional random field is defined as
\begin{equation}
L(\tilde{\theta}) = - \frac{1}{N} \sum_{i=1}^{N} \log L_i(\tilde{\theta}, M_i, y_i),
\end{equation}
where $\tilde{\theta} \in \mathbb{R}^K$ is the parameter we want to learn, and
\begin{equation}
L_i(\tilde{\theta}, M_i, y_i) = \tilde{\theta}^T M_i y_i - B(\tilde{\theta}^T M_i).
\label{eq:conditionallikelihood}
\end{equation}

The definition of $B(\tilde{\theta}^T M_i)$ is similar to that for MRFs. We can interpret $\tilde{\theta}^\top M_i$ to be the expanded, or ``grounded,'' potential vector. The product with feature matrix $M_i$ maps from the low-dimensional $\tilde{\theta}$ to a possibly high-dimensional, full potential vector $\theta$.
To be precise, the definition is
\begin{equation*}
B(\tilde{\theta}^T M_i) = \max_{\tau_i \in \mathbb{L}(G)} \{\tilde{\theta}^T M_i \tau_i - B^*(\tau_i)\}.
\end{equation*}

For each data point $i$, we can still use BBP to update $\tau_{i, t}$ at iteration $t$, i.e, lines 4--9 in Algorithm~\ref{PBP}. Let $\bar{w} = \frac{1}{N}\sum_{i}M_i y_i$. The approximate gradient is then
\begin{equation}
g(\tilde{\theta_t}) = \bar{w} - \frac{1}{N}\sum_{i=1}^{N}M_i \tau_{i,t}.
\end{equation}

Given $g(\tilde{\theta}_{t-1})$, we compute $g(\tilde{\theta}_{t})$ with the following rule:
\begin{equation*}
g(\tilde{\theta}_{t-1}) = g(\tilde{\theta}_{t-1}) + \frac{1}{N}\sum_{i=1}^{N} U_{i, F_t}(\tau_{i, t}^{(F_t)} - \tau_{i, t-1}^{(F_t)}).
\end{equation*}

Thus, the iteration complexity of inference only depends on the subnetwork size, rather than the size of the whole network. 

Regarding convergence, note that each matrix $M_i$ is constant, and its norm is bounded by its eigenvalues. Using the same proof method as that of Theorem \ref{theorem:converge}, we can straightforwardly prove that it still has a linear convergence rate.

The same formulation---using a matrix $M$ to map from a low-dimensional parameter vector $\tilde{\theta}$ to a possibly high-dimensional, full potential vector---can also be used to describe templated MRFs, where the same potential functions may be used in multiple parts of the graph. Therefore the same techniques and analysis also apply for templated MRFs.

\section{Empirical Study}
\label{sec:experiments}

In this section, we empirically analyze the performance of BBPL. We design two groups of experiments. In the first group of experiments, we evaluate the sensitivity of our method on the block size and empirically measure its convergence on synthetic networks. In the second group of experiments, we test our methods on a popular application of MRFs: image segmentation on a real image dataset.

\subsubsection{Baselines} We compare BBPL to other methods that exactly optimize the full variational likelihood. We use full convex BP and inner-dual learning~\cite{bach2015paired,hazan2016blending} as our baselines. Full BP learning runs inference to convergence each gradient step, and inner-dual learning accelerates learning by performing only one iteration of inference per learning iteration.

\subsubsection{Metrics} We evaluate the convergence and correctness of our method by measuring the objective value and the distance between the current parameter vector $\theta$ and the optimal $\theta^*$. To compute the objective value, we store the parameters $\theta_t$ obtained by full BP learning, BBPL, or inner-dual learning during training. We then plug each into Equation~\ref{eq:conditionallikelihood} to compute the variational negative log-likelihood. To compute the distance between $\theta_t$ and $\theta^*$, we first run convex BP learning until convergence to get $\theta^*$, and then we use the $\ell_2$ norm to measure the distance. 

\subsection{Experiments on Synthetic Networks}

\begin{figure}[tp]
	\centering
	\includegraphics[width= 0.48\textwidth]{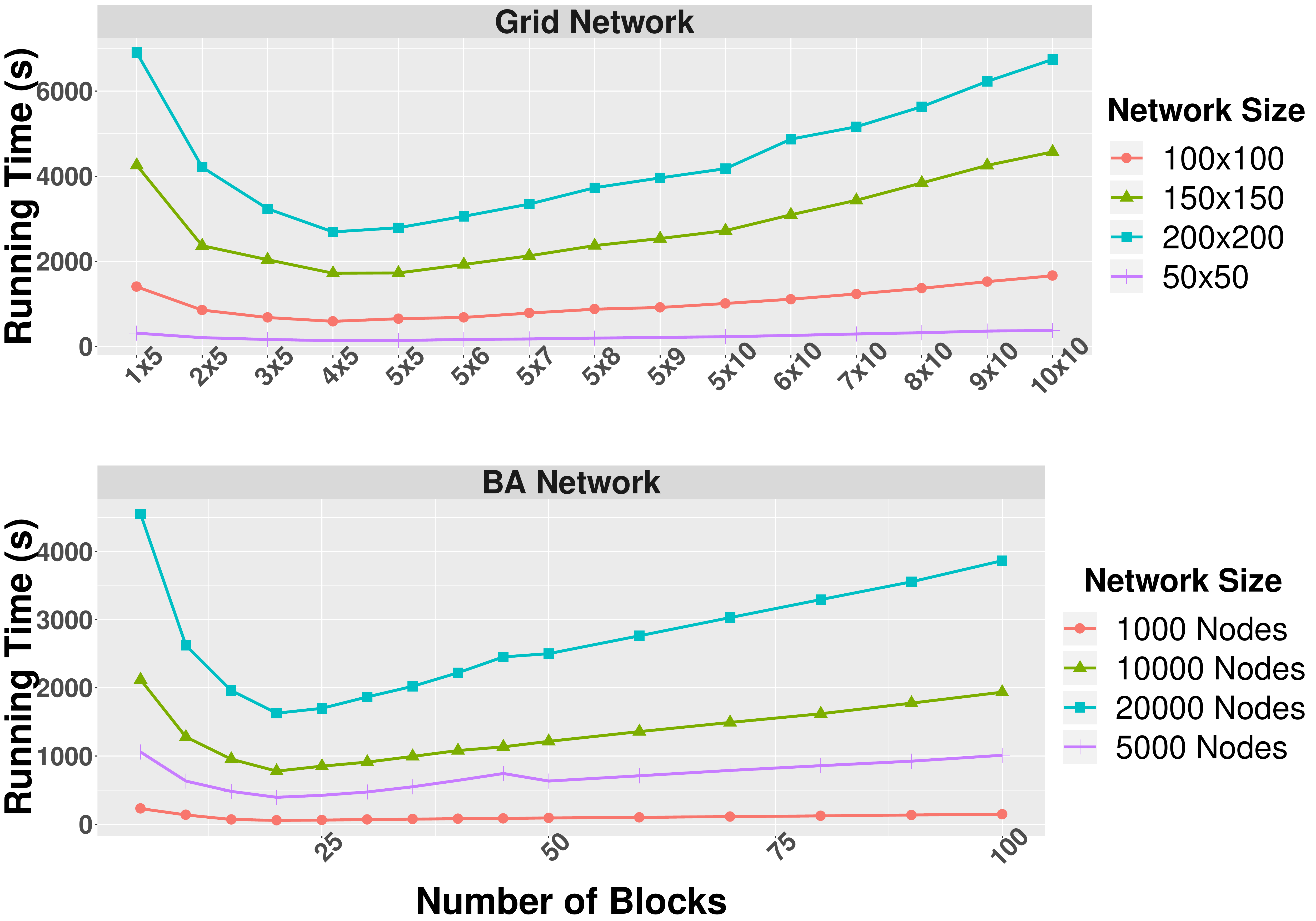}
	\caption{Sensitivity of BBPL to block size. The label $5 \times 5$ in the top polot represents that each grid is divided to 5 rows and 5 columns of sub-networks. The label $20$ in the bottom plot represents that the network is divided into 20 networks. The results show that BBPL converges the fastest when each block's size is about $20$ to $25$ times smaller than the network.}
	\label{fig:blocksize}       % Give a unique label
\end{figure}

We generate two types of synthetic MRFs: grid networks and Barabási–Albert (BA) random graph networks~\cite{albert2002statistical}. The nodes of the BA networks are ordered according to their generating sequence. We generate true unary and pairwise features from zero-mean, unit-variance Gaussian distributions. The unary feature dimension is 20 and the pairwise feature dimension is 10, and each variable has 8 possible states. Once the true models are generated, we draw 20 samples for each dataset with Gibbs sampling. 

\subsubsection{Sensitivity to block size} We conduct experiments on grid networks of four different sizes, i.e., $50 \times 50$, $100 \times 100$, $150 \times 150$, and $200 \times 200$, and on BA networks with four different sizes, i.e., 1,000 nodes, 5,000 nodes, 10,000 nodes, and 20,000 nodes. For the grid networks, we vary the number of blocks from $25$ $(5 \times 5)$ to $100$ $(10 \times 10)$. For example, when the network size is $200 \times 200$ and the number of blocks is $5 \times 5$, we separate the network into $25$ sub-networks, and each sub-network's size is $40 \times 40$. For the BA networks, we vary the number of blocks from $5$ to $100$. We separate the nodes of each BA network based on their indices. For example, for a node set $V = \{1,2,...,10\}$, if we want to separate it into two subsets, we will let $V_1 = {1,...,5}$, and $V_2 = \{6,...,10\}$.

The results are plotted in Figure~\ref{fig:blocksize}. The trends indicate that when each block is between $20$ to $25$ times smaller than the network, BBPL converges the fastest. When the block size is too small, the algorithm needs more iterations to converge. When the block size is too large, per-iteration complexity will be large. Blocks that are either too large or too small can reduce the benefits of BBPL.

\begin{figure*}[htp]
	\centering
	\includegraphics[width= 1\textwidth]{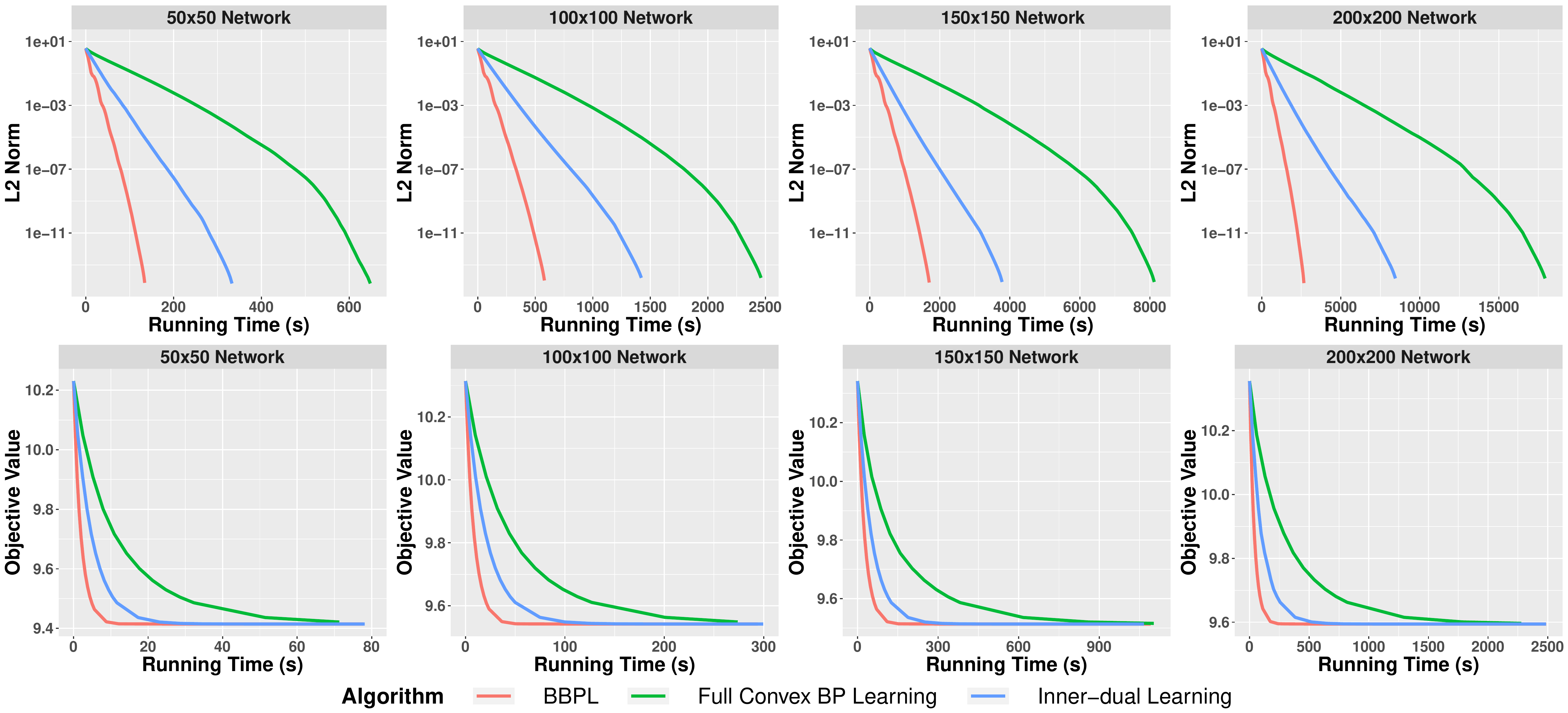}
	\caption{Convergence on grid networks. The top row plots the $\ell_2$ distance from the optimum, and the second row plots the objective values. The $\ell_2$ plots show the full running time of each method, but for clarity, we zoom into the plots of objective values by truncating the x-axis.}
	\label{fig:norm-obj}       % Give a unique label
\end{figure*}

\begin{figure*}[htp]
	\centering
	\includegraphics[width= 1\textwidth]{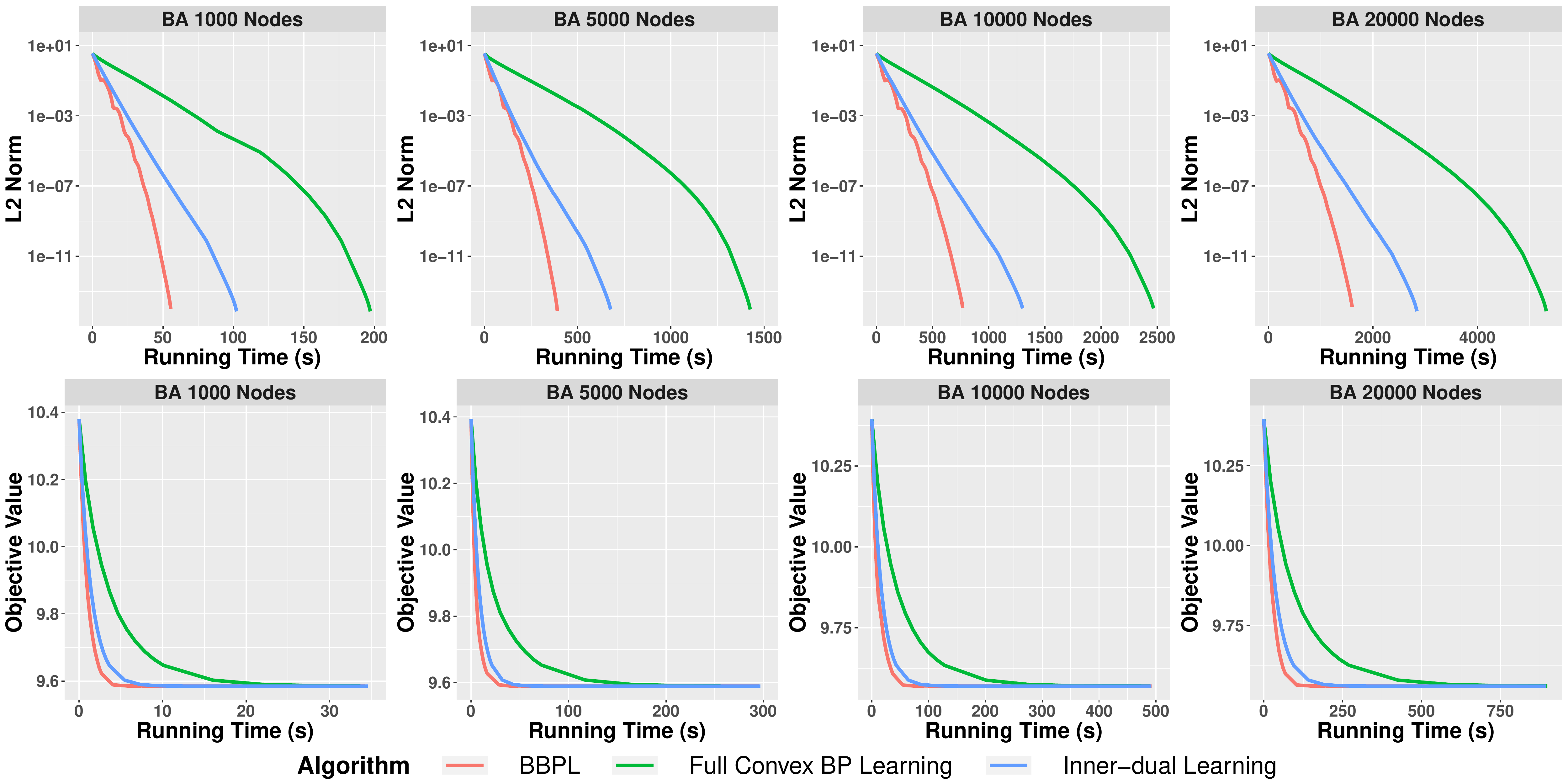}
	\caption{Results on BA networks. The top row again plots the $\ell_2$ distance from the optimum, and the bottom row plots the objective zoomed to show behavior early during optimization.}
	\label{fig:norm-obj-BA}       % Give a unique label
\end{figure*}

\subsubsection{Convergence analysis} 
\begin{figure}[!htp]
	\centering
	\includegraphics[width= 0.47\textwidth]{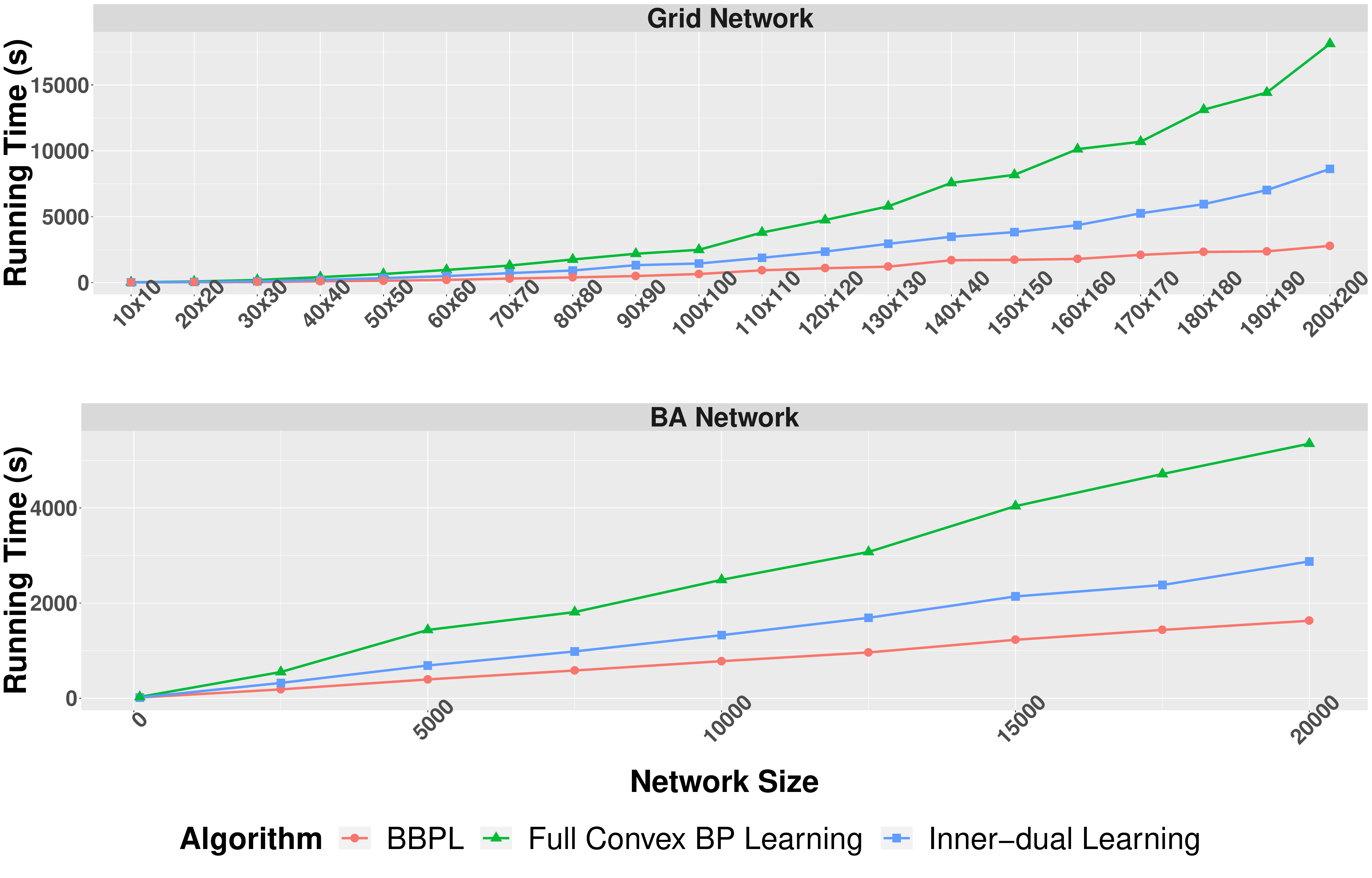}
	\caption{Comparisons of running time on networks with different sizes. BBPL converges much faster than the other two methods, especially on large networks where the improvement in running time is more significant.}
	\label{fig:runningtime}       % Give a unique label
\end{figure}

For the grid networks, we vary the network size from $10 \times 10$ to $200 \times 200$, and we set the number of blocks to $4 \times 5$. For the BA networks, we vary the network size from $100$ to 20,000, and we set the number of blocks to $20$. Figure~\ref{fig:norm-obj} and Figure~\ref{fig:norm-obj-BA} empirically show the convergence of BBPL on different networks. Figure~\ref{fig:runningtime}  plots the running time comparison on networks with all sizes. From Figure~\ref{fig:norm-obj} we can see that BBPL converges to the same solution as full BP learning, while requiring significantly less time. Figure~\ref{fig:runningtime} shows that on all networks, BBPL is the fastest. The acceleration is more significant when the network is large. For example, when the grid network's size is $200 \times 200$, BBPL is about three times faster than the inner-dual method and seven times faster than full convex BP. When the BA network size is 20,000, BBPL is about two times faster than inner-dual learning and three times faster than full convex BP.

\subsection{Experiments on Image Dataset}

For our real data experiments, we use the \emph{scene understanding} dataset~\cite{gould2009decomposing} for semantic image segmentation. Each image is $240 \times 320$ pixels in size. We randomly choose 50 images as the training set and 20 images as the test set. We extract unary features from a fully convolutional network (FCN)~\cite{long2015fully}. We add a linear transpose layer between the output layer and the last deconvolution layer of the FCN. This transpose layer does not impact the FCN's performance. Let $x \in \mathbb{R}^n$ be the output of the deconvolution layer and $y \in \mathbb{R}^c$ be the number of classes of the dataset. Then the input of the transpose layer is $x$ and its output is $y$. We use $x$ as the MRF's unary features. We use FCN-32 models to generate the features and we fine-tuned its parameters from the pre-trained VGG 16-layer network~\cite{simonyan2014very}. Our pairwise features are based on those of \citeauthor{domke2013learning} (\citeyear{domke2013learning}): for edge features of vertex $s$ and $t$, we compute the $\ell_2$ norm of unary features between $s$ and $t$ and discretize it to $10$ bins.

\begin{figure}[!htp]
	\centering
	\includegraphics[width= 0.47\textwidth]{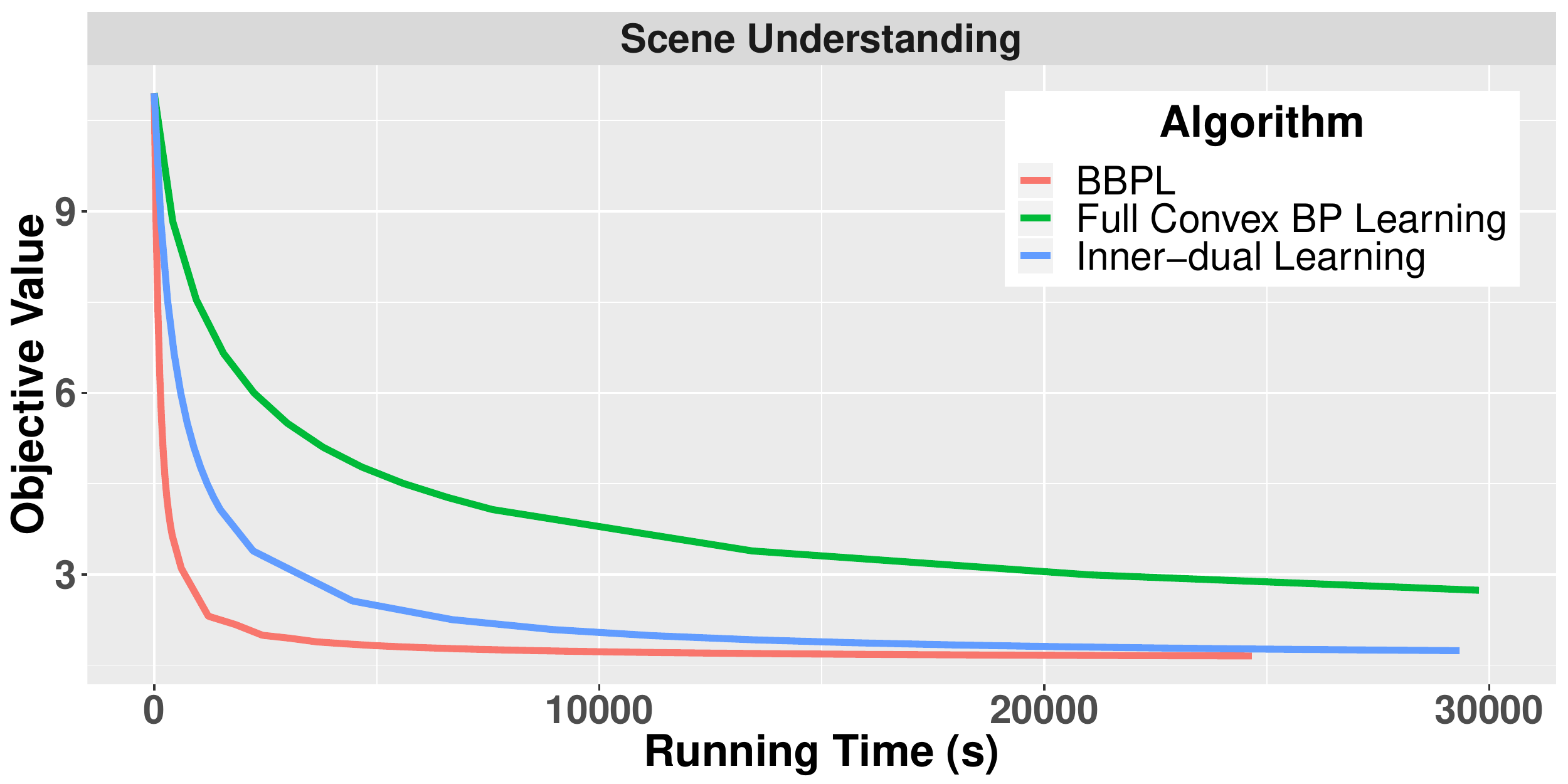}
	\caption{The learning objective during training on the scene understanding dataset. We run the three methods for the same amount of time. Our BBPL method is much faster than the two other methods.}
	\label{fig:scene-obj}       % Give a unique label
\end{figure}

We train MRFs on this segmentation task. Since these MRFs are large, full BP learning cannot run in a reasonable amount of time. Instead, we first run BBPL until convergence, and then we run inner-dual learning and full BP learning for the same amount of time. Finally, we compare their objective values during optimization. This evaluation scheme was necessary because the baseline methods need several days to converge on these large networks. The results are plotted in Figure~\ref{fig:scene-obj}, and Table~\ref{tab:table1} shows the number of iterations each algorithm runs when it stops. BBPL's per-iteration complexity is much lower than the other methods, so it runs the most number of iterations in the same time. Following the trends seen in the synthetic experiments, BBPL again reduces the objective much faster than the two other methods.

\begin{table}[!htp]
	\begin{center}
		\caption{Number of iterations each algorithm runs.}
		\label{tab:table1}
		\begin{tabular}{lll}
			\toprule
			\textbf{BBPL} & \textbf{Inner-dual Learning} & \textbf{Full BP Learning}\\
			\midrule
			601 & 163 & 21 \\
			\bottomrule
		\end{tabular}
	\end{center}
\end{table}

\section{Conclusion}
\label{sec:conclusion}

In this paper, we developed block belief propagation learning (BBPL), for training Markov random fields. At each learning iteration, our method only performs inference on a subnetwork, and it uses an approximation of the true gradient to optimize the parameters of interest. Thus, BBPL's iteration complexity does not scale with the size of the network. We theoretically prove that BBPL has a linear convergence rate and that it converges to the same optimum as convex BP. Our experiments show that, since BBPL has much lower iteration complexity, it converges faster than other methods that run (truncated or complete) inference on the full MRF each learning iteration.

For future work, we plan to use the scalability of BBPL to analyze large-scale networks. Further speedups may be possible. Even though BBPL only needs to run inference on a subnetwork, it still needs to run many iterations of belief propagation until convergence. We plan to develop a more efficient learning method that can stop inference in a fixed number of iterations, combining the benefits of BBPL and inner-dual learning. Finally, our proof depends on Assumption~\ref{assumption:contraction}, which has been made in the literature about belief propagation-like algorithms but has not been proven. We aim to both prove its validity for existing methods and use it to help derive new inference methods suitable for learning.

\section*{Acknowledgments}
\label{sec:acknowledgments}

We thank the anonymous reviewers, and Lijun Chen for for their insightful comments.
	
\fontsize{9.4pt}{10.5pt} \selectfont

\bibliographystyle{aaai}
	%\bibliographystyle{style/nips}
%\bibliography{youlu}

\begin{thebibliography}{}
	
	\bibitem[Ahmadi, Kersting, and
	Natarajan]{ahmadi2012lifted}
	Ahmadi, B.; Kersting, K.; and Natarajan, S.
	\newblock 2012.
	\newblock Lifted online training of relational models with stochastic gradient
	methods.
	\newblock In {\em Joint European Conference on Machine Learning and Knowledge
		Discovery in Databases}.
	
	\bibitem[\protect\citeauthoryear{Albert and
		Barab{\'a}si}{2002}]{albert2002statistical}
	Albert, R., and Barab{\'a}si, A.-L.
	\newblock 2002.
	\newblock Statistical mechanics of complex networks.
	\newblock {\em Reviews of Modern Physics} 74:47.
	
	\bibitem[\protect\citeauthoryear{Bach \bgroup et al\mbox.\egroup
	}{2015}]{bach2015paired}
	Bach, S.; Huang, B.; Boyd-Graber, J.; and Getoor, L.
	\newblock 2015.
	\newblock Paired-dual learning for fast training of latent variable hinge-loss
	{MRF}s.
	\newblock In {\em International Conference on Machine Learning}.
	
	\bibitem[\protect\citeauthoryear{Bixler and Huang}{2018}]{bixler2018sparse}
	Bixler, R., and Huang, B.
	\newblock 2018.
	\newblock Sparse matrix belief propagation.
	\newblock In {\em Conference on Uncertainty in Artificial Intelligence}.
	
	\bibitem[\protect\citeauthoryear{Bubeck}{2015}]{bubeck2015convex}
	Bubeck, S.
	\newblock 2015.
	\newblock Convex optimization: Algorithms and complexity.
	\newblock {\em Foundations and Trends{\textregistered} in Machine Learning}
	8:231--357.
	
	\bibitem[\protect\citeauthoryear{Domke}{2011}]{domke2011parameter}
	Domke, J.
	\newblock 2011.
	\newblock Parameter learning with truncated message-passing.
	\newblock In {\em Computer Vision and Pattern Recognition}.
	
	\bibitem[\protect\citeauthoryear{Domke}{2013}]{domke2013learning}
	Domke, J.
	\newblock 2013.
	\newblock Learning graphical model parameters with approximate marginal
	inference.
	\newblock {\em IEEE Transactions on Pattern Analysis and Machine Intelligence}
	35(10):2454--2467.
	
	\bibitem[\protect\citeauthoryear{Du and Hu}{2018}]{du2018linear}
	Du, S.~S., and Hu, W.
	\newblock 2018.
	\newblock Linear convergence of the primal-dual gradient method for
	convex-concave saddle point problems without strong convexity.
	\newblock {\em arXiv preprint arXiv:1802.01504}.
	
	\bibitem[\protect\citeauthoryear{Globerson and
		Jaakkola}{2007}]{globerson2007approximate}
	Globerson, A., and Jaakkola, T.
	\newblock 2007.
	\newblock Approximate inference using conditional entropy decompositions.
	\newblock In {\em Artificial Intelligence and Statistics}.
	
	\bibitem[\protect\citeauthoryear{Gould, Fulton, and
		Koller}{2009}]{gould2009decomposing}
	Gould, S.; Fulton, R.; and Koller, D.
	\newblock 2009.
	\newblock Decomposing a scene into geometric and semantically consistent
	regions.
	\newblock In {\em International Conference on Computer Vision}.
	
	\bibitem[\protect\citeauthoryear{Hazan and Urtasun}{2010}]{hazan2010primal}
	Hazan, T., and Urtasun, R.
	\newblock 2010.
	\newblock A primal-dual message-passing algorithm for approximated large scale
	structured prediction.
	\newblock In {\em Advances in Neural Information Processing Systems}.
	
	\bibitem[\protect\citeauthoryear{Hazan, Schwing, and
		Urtasun}{2016}]{hazan2016blending}
	Hazan, T.; Schwing, A.~G.; and Urtasun, R.
	\newblock 2016.
	\newblock Blending learning and inference in conditional random fields.
	\newblock {\em The Journal of Machine Learning Research} 17:8305--8329.
	
	\bibitem[\protect\citeauthoryear{Heskes}{2006}]{heskes2006convexity}
	Heskes, T.
	\newblock 2006.
	\newblock Convexity arguments for efficient minimization of the {B}ethe and
	{K}ikuchi free energies.
	\newblock {\em Journal of Artificial Intelligence Research} 26:153--190.
	
	\bibitem[\protect\citeauthoryear{Kakade, Shalev-Shwartz, and
		Tewari}{2009}]{kakade2009duality}
	Kakade, S.; Shalev-Shwartz, S.; and Tewari, A.
	\newblock 2009.
	\newblock On the duality of strong convexity and strong smoothness: Learning
	applications and matrix regularization.
	\newblock Technical report.
	
	\bibitem[\protect\citeauthoryear{Kersting, Ahmadi, and
		Natarajan}{2009}]{kersting2009counting}
	Kersting, K.; Ahmadi, B.; and Natarajan, S.
	\newblock 2009.
	\newblock Counting belief propagation.
	\newblock In {\em Uncertainty in Artificial Intelligence}.
	
	\bibitem[\protect\citeauthoryear{Koller and
		Friedman}{2009}]{koller2009probabilistic}
	Koller, D., and Friedman, N.
	\newblock 2009.
	\newblock {\em Probabilistic Graphical Models: Principles and Techniques}.
	\newblock MIT press.
	
	\bibitem[\protect\citeauthoryear{Lin \bgroup et al\mbox.\egroup
	}{2015}]{lin2015deeply}
	Lin, G.; Shen, C.; Reid, I.; and van~den Hengel, A.
	\newblock 2015.
	\newblock Deeply learning the messages in message passing inference.
	\newblock In {\em Advances in Neural Information Processing Systems}.
	
	\bibitem[\protect\citeauthoryear{London, Huang, and
		Getoor}{2015}]{london2015benefits}
	London, B.; Huang, B.; and Getoor, L.
	\newblock 2015.
	\newblock The benefits of learning with strongly convex approximate inference.
	\newblock In {\em International Conference on Machine Learning}.
	
	\bibitem[\protect\citeauthoryear{Long, Shelhamer, and
		Darrell}{2015}]{long2015fully}
	Long, J.; Shelhamer, E.; and Darrell, T.
	\newblock 2015.
	\newblock Fully convolutional networks for semantic segmentation.
	\newblock In {\em Computer Vision and Pattern Recognition},  3431--3440.
	
	\bibitem[\protect\citeauthoryear{Meshi \bgroup et al\mbox.\egroup
	}{2009}]{meshi2009convexifying}
	Meshi, O.; Jaimovich, A.; Globerson, A.; and Friedman, N.
	\newblock 2009.
	\newblock Convexifying the {B}ethe free energy.
	\newblock In {\em Conference on Uncertainty in Artificial Intelligence}.
	
	\bibitem[\protect\citeauthoryear{Meshi \bgroup et al\mbox.\egroup
	}{2010}]{meshi2010learning}
	Meshi, O.; Sontag, D.; Jaakkola, T.; and Globerson, A.
	\newblock 2010.
	\newblock Learning efficiently with approximate inference via dual losses.
	\newblock In {\em International Conference on Machine Learning}.
	
	\bibitem[\protect\citeauthoryear{Noorshams and
		Wainwright}{2013}]{noorshams2013stochastic}
	Noorshams, N., and Wainwright, M.~J.
	\newblock 2013.
	\newblock Stochastic belief propagation: A low-complexity alternative to the
	sum-product algorithm.
	\newblock {\em IEEE Transactions on Information Theory} 59:1981--2000.
	
	\bibitem[\protect\citeauthoryear{Nowozin and
		Lampert}{2011}]{nowozin2011structured}
	Nowozin, S., and Lampert, C.~H.
	\newblock 2011.
	\newblock Structured learning and prediction in computer vision.
	\newblock {\em Foundations and Trends in Computer Graphics and Vision}
	6:185--365.
	
	\bibitem[\protect\citeauthoryear{Richardson and
		Domingos}{2006}]{richardson2006markov}
	Richardson, M., and Domingos, P.
	\newblock 2006.
	\newblock Markov logic networks.
	\newblock {\em Machine Learning} 62:107--136.
	
	\bibitem[\protect\citeauthoryear{Roosta, Wainwright, and
		Sastry}{2008}]{roosta2008convergence}
	Roosta, T.~G.; Wainwright, M.~J.; and Sastry, S.~S.
	\newblock 2008.
	\newblock Convergence analysis of reweighted sum-product algorithms.
	\newblock {\em IEEE Transactions on Signal Processing} 56:4293--4305.
	
	\bibitem[\protect\citeauthoryear{Ross \bgroup et al\mbox.\egroup
	}{2011}]{ross2011learning}
	Ross, S.; Munoz, D.; Hebert, M.; and Bagnell, J.~A.
	\newblock 2011.
	\newblock Learning message-passing inference machines for structured
	prediction.
	\newblock In {\em Computer Vision and Pattern Recognition}.
	
	\bibitem[\protect\citeauthoryear{Samdani and Roth}{2012}]{samdani2012efficient}
	Samdani, R., and Roth, D.
	\newblock 2012.
	\newblock Efficient decomposed learning for structured prediction.
	\newblock {\em arXiv preprint arXiv:1206.4630}.
	
	\bibitem[\protect\citeauthoryear{Schwing \bgroup et al\mbox.\egroup
	}{2011}]{schwing2011distributed}
	Schwing, A.; Hazan, T.; Pollefeys, M.; and Urtasun, R.
	\newblock 2011.
	\newblock Distributed message passing for large scale graphical models.
	\newblock In {\em Computer Vision and Pattern Recognition}.
	
	\bibitem[\protect\citeauthoryear{Simonyan and
		Zisserman}{2014}]{simonyan2014very}
	Simonyan, K., and Zisserman, A.
	\newblock 2014.
	\newblock Very deep convolutional networks for large-scale image recognition.
	\newblock {\em arXiv preprint arXiv:1409.1556}.
	
	\bibitem[\protect\citeauthoryear{Singla and Domingos}{2008}]{singla2008lifted}
	Singla, P., and Domingos, P.~M.
	\newblock 2008.
	\newblock Lifted first-order belief propagation.
	\newblock In {\em Association for the Advancement of Artificial Intelligence}.
	
	\bibitem[\protect\citeauthoryear{Stoyanov, Ropson, and
		Eisner}{2011}]{stoyanov2011empirical}
	Stoyanov, V.; Ropson, A.; and Eisner, J.
	\newblock 2011.
	\newblock Empirical risk minimization of graphical model parameters given
	approximate inference, decoding, and model structure.
	\newblock In {\em International Conference on Artificial Intelligence and
		Statistics},  725--733.
	
	\bibitem[\protect\citeauthoryear{Sutton and McCallum}{2009}]{sutton08piecewise}
	Sutton, C., and McCallum, A.
	\newblock 2009.
	\newblock Piecewise training for structured prediction.
	\newblock {\em Machine Learning} 77(2--3):165--194.
	
	\bibitem[\protect\citeauthoryear{Taskar \bgroup et al\mbox.\egroup
	}{2005}]{taskar2005learning}
	Taskar, B.; Chatalbashev, V.; Koller, D.; and Guestrin, C.
	\newblock 2005.
	\newblock Learning structured prediction models: A large margin approach.
	\newblock In {\em International Conference on Machine Learning}.
	
	\bibitem[\protect\citeauthoryear{Taskar, Guestrin, and
		Koller}{2004}]{taskar2004max}
	Taskar, B.; Guestrin, C.; and Koller, D.
	\newblock 2004.
	\newblock Max-margin markov networks.
	\newblock In {\em Advances in Neural Information Processing Systems}.
	
	\bibitem[\protect\citeauthoryear{Wainwright and
		Jordan}{2008}]{wainwright2008graphical}
	Wainwright, M.~J., and Jordan, M.~I.
	\newblock 2008.
	\newblock Graphical models, exponential families, and variational inference.
	\newblock {\em Foundations and Trends in Machine Learning} 1:1--305.
	
	\bibitem[\protect\citeauthoryear{Wainwright, Jaakkola, and
		Willsky}{2005}]{wainwright2005new}
	Wainwright, M.~J.; Jaakkola, T.~S.; and Willsky, A.~S.
	\newblock 2005.
	\newblock A new class of upper bounds on the log partition function.
	\newblock {\em IEEE Transactions on Information Theory} 51:2313--2335.
	
	\bibitem[\protect\citeauthoryear{Wainwright}{2006}]{wainwright2006estimating}
	Wainwright, M.~J.
	\newblock 2006.
	\newblock Estimating the``wrong''graphical model: Benefits in the
	computation-limited setting.
	\newblock {\em Journal of Machine Learning Research} 7:1829--1859.
	
	\bibitem[\protect\citeauthoryear{Yedidia, Freeman, and
		Weiss}{2005}]{yedidia2005constructing}
	Yedidia, J.~S.; Freeman, W.~T.; and Weiss, Y.
	\newblock 2005.
	\newblock Constructing free-energy approximations and generalized belief
	propagation algorithms.
	\newblock {\em IEEE Transactions on Information Theory} 51:2282--2312.
	
	\bibitem[\protect\citeauthoryear{Yin and Gao}{2014}]{yin2014scalable}
	Yin, J., and Gao, L.
	\newblock 2014.
	\newblock Scalable distributed belief propagation with prioritized block
	updates.
	\newblock In {\em International Conference on Conference on Information and
		Knowledge Management}.
	
\end{thebibliography}

%\input{2019_aaai_block_BP.bbl}

\newpage
    
\fontsize{10pt}{11pt} \selectfont

\section{Proof of  Theorem 1}
\label{sec:proof}

The following two lemmas provide preliminaries for the proof.

\begin{lemma}
	The function 
	\begin{equation*}
	L(\theta) = -\theta^T \bar{w} + B(\theta)
	\end{equation*} 
	is $\beta$-strongly convex and $\eta$-smooth.
	\label{lemma:strongly_convex}
\end{lemma}

\textit{Proof.} The function $B^*(\tau)$ is strongly convex and smooth~\cite{meshi2009convexifying,yedidia2005constructing,wainwright2006estimating}. The function $B(\theta)$ is the conjugate function of $B^*(\tau)$. Based on the conjugate function's property \cite{kakade2009duality}, $B(\theta)$ is also strongly convex and smooth. The other term of $L(\theta)$ is the Euclidean inner product of $\theta$ and $\bar{w}$, where $\bar{w}$ is constant. Thus, $L(\theta)$ is also strongly convex and smooth. \hfill $\square$

\begin{lemma}
	Suppose $f: \mathbb{R^d} \to \mathbb{R}$ is a $\beta$-strongly convex and $\eta$-smooth function. Let $x^* = \arg \min f(x)$. When using gradient descent to optimize $x$, i.e., $x_{t+1} = x_{t} - \alpha \nabla f(x_t)$, with the learning rate $0 < \alpha \le \frac{2}{\beta + \eta}$, we have
	\begin{equation*}
	||x_{t+1} - x^*|| \le (1 - \beta \alpha)||x_{t} - x^*||.
	\end{equation*}
	\label{lemma:strongly_convex_convergence}
\end{lemma}

\textit{Proof.} See Theorem 3.12 of~\cite{bubeck2015convex}.
\hfill $\square$

The proof follows three steps, each detailed in its own section below.

\subsection{Step 1: Bounding the Decrease of $||\theta_t - \theta^*||$}

In the first step, we want to prove that $||\theta_{t+1} - \theta^*||$ is upper bounded by the weighted sum of $||\theta_{t} - \theta^*||$ and $||\tau_t - \tau^*_t||$.

\begin{lemma}
	Let $\hat{\theta}_{t+1} = \theta_t - \alpha_t \nabla L(\theta_t) $, where $0 < \alpha_t \le \frac{2}{\beta + \eta}$. Then
	\begin{equation*}
	||\hat{\theta}_{t+1} - \theta^*|| \le (1 - \beta \alpha_t) ||\theta_t - \theta^*||.
	\end{equation*}
	\label{proposition_theta_hat}
\end{lemma} 

\textit{Proof.} From Lemma~\ref{lemma:strongly_convex}, we know that $L(\theta)$ is $\beta$-strongly convex and $\eta$-smooth. The parameter vector $\hat{\theta}_{t+1}$ is obtained via one step of the gradient descent update. This lemma then follows from Lemma~\ref{lemma:strongly_convex_convergence}. \hfill $\square$

\begin{lemma}
	Let $\theta_{t+1} = \theta_{t} - \alpha_t g(\theta_t)$, where $0 < \alpha_t \le \frac{2}{\beta + \rho}$. Then
	\begin{equation*}
	||\theta_{t+1} - \theta^*|| \le (1-\beta \alpha_t)||\theta_t - \theta^*|| + \alpha_t ||\tau_t - \tau^*_t|| .
	\end{equation*}
	\label{proposition:theta_star}
\end{lemma}

\textit{Proof.} From the update rules of $\theta_{t+1}$ and $\hat{\theta}_{t+1}$, we have
\begin{eqnarray*}
	||\hat{\theta}_{t+1} - \theta_{t+1}|| &=& ||\alpha_t(g(\theta_t) - \nabla L(\theta_t))|| \\
	&=& ||\alpha_t(\tau_t - \tau_t^*)|| \\
	&\le& \alpha_t ||\tau_t - \tau_t^*||.
\end{eqnarray*}

Using Lemma~\ref{proposition_theta_hat} and the triangle inequality, we have
\begin{eqnarray*}
	||\theta_{t+1} - \theta^*|| &=& ||\theta_{t+1} - \hat{\theta}_{t+1} + \hat{\theta}_{t+1} -  \theta^*|| \\
	&\le&||\theta_{t+1} - \hat{\theta}_{t+1} || + ||\hat{\theta}_{t+1} -  \theta^*|| \\
	&\le& (1 - \beta \alpha_t) ||\theta_t - \theta^*|| + \alpha_t ||\tau_t - \tau_t^*||.
\end{eqnarray*}

The last inequality proves the lemma. \hfill $\square$

\subsection{Step 2: Bounding the Decrease of $||\tau_t - \tau_t^*||$}

In the second step, we want to prove that the $||\tau_{t+1} - \tau^*||$ is upper bounded by the weighted sum of $||\tau_t - \tau^*_t||$ and $||\theta_{t} - \theta^*||$.

\begin{lemma}
	Based on the update rules of $\tau$ and $\theta$, we have
	\begin{equation*}
	||\tau_{t+1}^* - \tau_t^*|| \le \eta \alpha_t ||\tau_t - \tau^*_t|| + \eta^2 \alpha_t ||\theta_t - \theta^*||.
	\end{equation*}
	\label{proposition:difference}
\end{lemma}

\textit{Proof.}
\begin{eqnarray*}
	||\tau_{t+1}^* - \tau_t^*|| &\le& \eta||\theta_{t+1} - \theta_{t}|| \\
	&=& \eta \alpha_t  ||-\bar{w} + \tau_t|| \\
	&\le& \eta \alpha_t (||-\bar{w} + \tau_t^*|| + ||\tau_t - \tau^*_t||) \\
	&=& \eta \alpha_t (||\nabla L(\theta_t) - \nabla L(\theta^*)|| \\
	&\quad& ~~~~~~~ + ||\tau_t - \tau^*_t||) \\
	&\le&\eta^2 \alpha_t ||\theta_t - \theta^*|| + \eta \alpha_t ||\tau_t - \tau^*_t||.
\end{eqnarray*}

In this proof, we use Lemma~\ref{lemma:strongly_convex}, which shows that $L(\theta)$ is $\eta$-smooth, and the gradient $\nabla L(\theta_t) = -\bar{w} + \tau^*_t$. For the second equation, we use the fact that $\nabla L(\theta^*) = 0$. \hfill $\square$

\begin{lemma}
	Suppose that block convex BP satisfies Assumption 1. Then 
	\begin{eqnarray*}
		||\tau_{t+1} - \tau_{t+1}^*|| &\le& (1 - c + \eta \alpha_t - c\eta \alpha_t)||\tau_t - \tau^*_t||\\
		&\quad& + (\eta^2 \alpha_t - c \eta^2 \alpha_t) ||\theta_t - \theta^*||,
	\end{eqnarray*}
	\label{proposition:tau_star}
	where $0 < c < 1$.
\end{lemma}

\textit{Proof.} 
\begin{eqnarray*}
	||\tau_{t+1} - \tau_{t+1}^*|| &\le& (1-c)||\tau_{t} - \tau_{t+1}^*||  \\
	&\le& (1-c) (||\tau_t - \tau^*_t|| + ||\tau_{t+1}^* - \tau_{t}^*||) \\
	&\le& (1 - c)(||\tau_t - \tau^*_t|| ) \\
	&\quad& + (1-c)(\eta \alpha_t ||\tau_t - \tau^*_t||) \\
	&\quad& + (1-c)(\eta^2 \alpha_t ||\theta_t - \theta^*||) \\
	&=& (1 - c + \eta \alpha_t - c \eta \alpha_t)||\tau_t - \tau_t^*|| \\
	&\quad& + (\eta^2 \alpha_t - c \eta^2 \alpha_t) ||\theta_t - \theta^*||.
\end{eqnarray*}

In this proof, we first use Assumption 1 to upper bound $||\tau_{t+1} - \tau_{t+1}^*||$, and then we use Lemma~\ref{proposition:difference} to expand the term $||\tau^*_{t+1} - \tau^*_t||$. \hfill $\square$

\subsection{Step 3: Proving the Decrease}

With the lemmas above, we can prove that $P_{t+1} \le (1-\delta) P_t$. 

Based on Lemma~\ref{proposition:theta_star} and Lemma~\ref{proposition:tau_star}, we have
\begin{eqnarray*}
	P_{t+1} &=& ||\theta_{t+1} - \theta^*|| + \gamma ||\tau_{t+1} - \tau_{t+1}^*|| \\
	&\le& (1-\beta \alpha_t)||\theta_t - \theta^*|| + \alpha_t ||\tau_t - \tau^*_t|| \\
	&\quad& + \gamma (1 - c + \eta \alpha_t - c \eta \alpha_t)||\tau_t - \tau_t^*||\\
	&\quad& + \gamma (\eta^2 \alpha_t - c \eta^2 \alpha_t) ||\theta_t - \theta^*||   \\
	&=& (1 - \beta \alpha_t + \gamma (\eta^2 \alpha_t - c \eta^2 \alpha_t)) ||\theta_t - \theta^*|| \\
	&\quad& + (\alpha_t + \gamma (1 - c + \eta \alpha_t - c \eta \alpha_t)) ||\tau_t - \tau^*_t|| \\
	&=& (1 - \beta \alpha_t + \gamma (\eta^2 \alpha_t - c \eta^2 \alpha_t))(||\theta_t - \theta^*|| \\
	&\quad& + \frac{\alpha_t + \gamma (1 - c + \eta \alpha_t - c \eta \alpha_t)}{1 - \beta \alpha_t + \gamma (\eta^2 \alpha_t - c \eta^2 \alpha_t)}||\tau_t - \tau^*_t||).
\end{eqnarray*}

For BBPL to be linearly convergent, we need to have
\begin{equation}
1 - \beta \alpha_t + \gamma (\eta^2 \alpha_t - c \eta^2 \alpha_t) < 1
\label{eq:goal1}
\end{equation}
and 
\begin{equation}
\frac{\alpha_t + \gamma (1 - c + \eta \alpha_t - c \eta \alpha_t)}{1 - \beta \alpha_t + \gamma (\eta^2 \alpha_t - c \eta^2 \alpha_t)} \le \gamma .
\label{eq:goal2}
\end{equation}

From Equation~\ref{eq:goal1}, we need to satisfy
\begin{equation}
\gamma < \frac{\beta}{(1-c)\eta^2}.
\label{eq:gammaupbound1}
\end{equation}

We expand Equation~\ref{eq:goal2} to obtain
\begin{eqnarray}
&& \alpha_t + \gamma (1 - c + \eta \alpha_t - c \eta \alpha_t) \nonumber\\
&\quad& ~~~~~~~ \le \gamma (1 - \beta \alpha_t + \gamma (\eta^2 \alpha_t - c \eta^2 \alpha_t)) \nonumber\\[8pt]
&\Rightarrow& \alpha_t + \gamma \eta \alpha_t - c \gamma \eta \alpha_t - c\gamma \nonumber\\
&\quad&~~~~~~~\le -\beta \gamma \alpha_t + \gamma^2 \eta^2 \alpha_t - c^2 \eta^2 \gamma^2 \alpha_t \nonumber\\[8pt]
&\Rightarrow& \eta^2 \alpha_t(1-c^2)\gamma^2 + (c \eta \alpha_t - \eta \alpha_t + c - \beta \alpha_t)\gamma \nonumber\\
&\quad&~~~~~~~ - \alpha_t \ge 0.
\label{eq:expandgoal2}
\end{eqnarray}

Note that the Equation~\ref{eq:expandgoal2} is a quadratic function of $\gamma$, with $\eta^2 \alpha_t(1-c^2)>0$ and $-\alpha_t <0$. Thus, it is a convex function, and has one positive root, i.e., $\gamma_1$, and one negative root, i.e., $\gamma_2$. To satisfy both Equations~\ref{eq:gammaupbound1} and \ref{eq:expandgoal2}, we need that $\gamma_1 < \gamma < \frac{\beta}{(1-c)\eta^2}$. Now we set

\begin{equation}
\gamma = \frac{\beta}{2(1-c)\eta^2}.
\label{eq:gammavalue}
\end{equation}

To make the chosen $\gamma$ lie in the interval above requires that when $\gamma = \frac{\beta}{2(1-c)\eta^2}$, the Equation~\ref{eq:expandgoal2} is satisfied.

Substituting Equation~\ref{eq:gammavalue} into Equation~\ref{eq:expandgoal2}, we have that

\begin{eqnarray}
&&\eta^2 \alpha_t(1-c^2) \frac{\beta^2}{4(1-c)^2\eta^4} \nonumber\\
&\quad&~~~~~~~+ \frac{\beta}{2(1-c)\eta^2}(c \eta \alpha_t - \eta \alpha_t + c - \beta \alpha_t) - \alpha_t \ge 0 \nonumber\\[8pt]
&\Rightarrow& \frac{(1-c^2)\beta^2}{4(1-c)^2 \eta^2} \nonumber\\
&\quad&~~~~~~~ + \frac{\beta}{2(1-c)\eta^2}(c\eta -\eta + \frac{c}{\alpha_t} - \beta) - 1 \ge 0 \nonumber \\[8pt]
&\Rightarrow& \frac{(1+c)\beta^2}{4(1-c)\eta^2} - \frac{c \beta }{2\eta} \nonumber\\
&\quad&~~~~~~~+ \frac{c \beta}{2 \alpha_t (1-c)\eta^2} - \frac{\beta^2}{2(1-c)\eta^2} - 1 \ge 0 \nonumber\\[8pt]
&\Rightarrow& \frac{c \beta}{\alpha_t (1-c)\eta^2} \ge 2 + \frac{c \beta}{\eta} + \frac{\beta^2}{2\eta^2}.
\label{eq:alphainequality}
\end{eqnarray}

In the first arrow, we divide both sides of the inequality by $\alpha_t$. 

Note that $\frac{c \beta}{\alpha_t (1-c)\eta^2} \ge \frac{c \beta}{\alpha_t \eta^2}$, and $2+\frac{\beta}{\eta} + \frac{\beta^2}{\eta^2} \ge 2 + \frac{c \beta}{\eta} + \frac{\beta^2}{2\eta^2}$. To satisfy Equation~\ref{eq:alphainequality}, we can require that 

\begin{eqnarray}
\frac{c \beta}{\alpha_t \eta^2} \ge 2 + \frac{\beta}{\eta} + \frac{\beta^2}{\eta^2}.
\label{eq:alphainequalityfinal}
\end{eqnarray}

Equation~\ref{eq:alphainequalityfinal} is equivalent to the condition
\begin{equation}
\alpha_t \le \frac{c \beta}{2\eta^2 + \eta \beta + \beta^2}.
\label{eq:alphabound}
\end{equation}

Let 
\begin{eqnarray*}
	\delta &=& \beta \alpha_t - \gamma (\eta^2 \alpha_t - c \eta^2 \alpha_t) \\
	&=& \beta \alpha_t - \frac{\beta}{2(1-c)\eta^2}(\eta^2 \alpha_t - c \eta^2 \alpha_t) \\
	&=& \frac{\beta\alpha_t}{2}.
\end{eqnarray*}
It is easy to prove that $0 < \delta < 1$, when $\alpha_t$ satisfies Equation~\ref{eq:alphabound}. Thus, we have that 
\begin{equation}
P_{t+1} \le (1- \delta) P_t,
\end{equation}
which implies linear convergence and proves the theorem. \hfill $\square$ 

\section{Sample Segmentation Results}

\begin{figure}[!htp]
	\subfloat
	{\includegraphics[width=.3\linewidth]{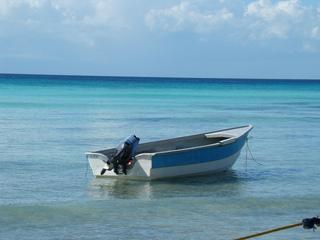}}\hfill
	\subfloat
	{\includegraphics[width=.3\linewidth]{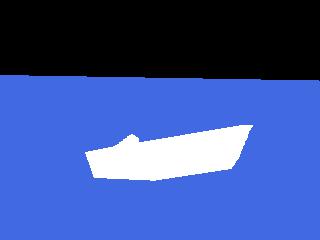}}\hfill
	\subfloat
	{\includegraphics[width=.3\linewidth]{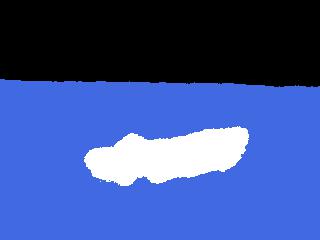}} \hfill
	{\includegraphics[width=.3\linewidth]{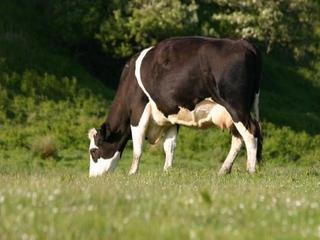}}\hfill
	\subfloat
	{\includegraphics[width=.3\linewidth]{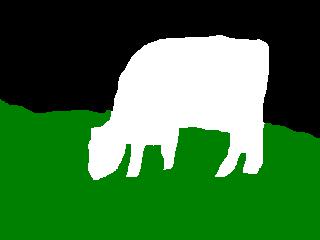}}\hfill
	\subfloat
	{\includegraphics[width=.3\linewidth]{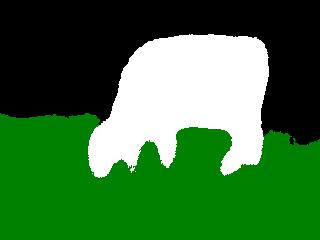}} \hfill
	{\includegraphics[width=.3\linewidth]{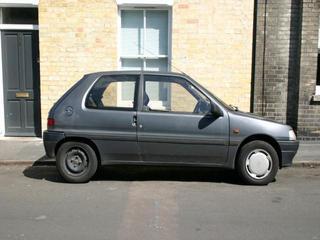}}\hfill
	\subfloat
	{\includegraphics[width=.3\linewidth]{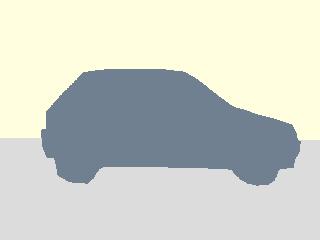}}\hfill
	\subfloat
	{\includegraphics[width=.3\linewidth]{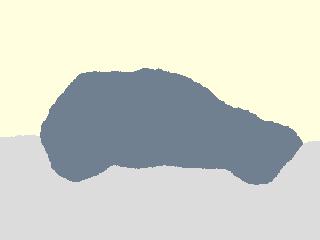}} \hfill
	\caption{Example segmentation results on the Scene Understanding dataset. From left to right: the original image, the ground-truth labels, and the predictions made by the optimized models.}
	\label{fig:scene}
\end{figure}

Since our method converges to the same optimum as the convex BP, these two methods have the same segmentation results. The pixel accuracy is $80.4\%$ on the Scene Understanding dataset. Figure~\ref{fig:scene} contains some examples of segmentation results. The images are selected from the test set.

\FloatBarrier

\end{document}